\documentclass[conference]{IEEEtran}
\IEEEoverridecommandlockouts
\usepackage{cite}
\usepackage{amsmath,amssymb,amsfonts}
\usepackage{algorithmic}
\usepackage{graphicx}
\usepackage{textcomp}
\usepackage{xcolor}
\usepackage[a4paper, total={184mm,239mm}]{geometry}
\def\BibTeX{{\rm B\kern-.05em{\sc i\kern-.025em b}\kern-.08em
		T\kern-.1667em\lower.7ex\hbox{E}\kern-.125emX}}

\usepackage[utf8]{inputenc}
\usepackage{mathtools}
\usepackage{amsthm, color}
\usepackage{bm, makecell}
\usepackage{tikz, subcaption, cite}
\usepackage{multirow}
\usepackage{array,booktabs,arydshln,xcolor}
\usepackage{algorithm}
\usepackage{flushend}
\graphicspath{{fig/}}
\DeclareMathOperator*{\argmin}{argmin}

\begin{document}

\title{Experts in the Loop: Conditional Variable Selection for Accelerating Post-Silicon Analysis Based on Deep Learning\thanks{This research was supported by Advantest as part of the Graduate School ``Intelligent Methods for Test and Reliability'' (GS-IMTR) at the University of Stuttgart.}}

\author{\IEEEauthorblockN{Yiwen Liao\IEEEauthorrefmark{1}, Rapha\"el Latty\IEEEauthorrefmark{2}, and Bin Yang\IEEEauthorrefmark{1}}
	\IEEEauthorblockA{\IEEEauthorrefmark{1}Institute of Signal Processing and System Theory, University of Stuttgart, Germany}
	\IEEEauthorblockA{\IEEEauthorrefmark{2}Applied Research and Venture Team, Advantest Europe GmbH, Germany}
}

\maketitle

\begin{abstract}
	Post-silicon validation is one of the most critical processes in modern semiconductor manufacturing. Specifically, correct and deep understanding in test cases of manufactured devices is key to enable post-silicon tuning and debugging. This analysis is typically performed by experienced human experts. However, with the fast development in semiconductor industry, test cases can contain hundreds of variables. The resulting high-dimensionality poses enormous challenges to experts. Thereby, some recent prior works have introduced data-driven variable selection algorithms to tackle these problems and achieved notable success. Nevertheless, for these methods, experts are not involved in training and inference phases, which may lead to bias and inaccuracy due to the lack of prior knowledge. Hence, this work for the first time aims to design a novel conditional variable selection approach while keeping experts in the loop. In this way, we expect that our algorithm can be more efficiently and effectively trained to identify the most critical variables under certain expert knowledge. Extensive experiments on both synthetic and real-world datasets from industry have been conducted and shown the effectiveness of our method.
\end{abstract}


\section{Introduction}
\label{sec:introduction}
The past decades have seen increasingly rapid advances in the semiconductor industry. One of the main motivations behind is that semiconductors are everywhere in human's daily activities, including but not limited to smart phones, personal computers, autonomous vehicles, medical equipment and cloud services. The resulting enormous demand on chip-based devices poses great challenges to semiconductor manufacturing, from design phases, pre- and post-silicon validation, to volume and in-field test. Among different phases within the manufacturing, Post-Silicon Validation (PSV) is known as one of the most challenging and costly components of the entire validation procedure for chip design as stated in~\cite{mishra2019post,7892969}.

\begin{figure}
	\centering
	\begin{tikzpicture}[scale=1]
		\definecolor{mycolor1}{RGB}{254,224,144}  
		\definecolor{mycolor2}{RGB}{217,239,139} 
		\definecolor{mycolor3}{RGB}{178,171,210} 

		\draw[thick, black!55, rounded corners=1mm,fill=mycolor2, fill opacity=0.5] (-0., -1.) -- (-0., 1.) -- (0.5, 1.) -- (0.5, -1.) -- cycle;
		\node (c1) at(0.25, .75) {$c_1$};
		\node[anchor=mid] (c2) at(0.25, 0) {$\vdots$};
		\node (c3) at(0.25, -.75) {$c_N$};

		\draw[thick, black!55, rounded corners=1mm,fill=blue!15, fill opacity=0.5] (1.5, -0.5) -- (1.5, 0.5) -- (3.5, 0.5) -- (3.5, -0.5) -- cycle;
		\node (dut) at(2.5, 0) {DUT};
		
		\draw[thick, black!55, rounded corners=1mm,fill=mycolor1, fill opacity=0.5] (1.25, 1) -- (1.25, 1.5) -- (3.75, 1.5) -- (3.75, 1) -- cycle;
		\node (t1) at(1.5, 1.25) {$t_1$};
		\node (t2) at(2.5, 1.25) {$\dots$};
		\node (t3) at(3.5, 1.25) {$t_M$};
		
		\draw[thick, black!55,->] (.5, 0)--(1.5, 0);
		\draw[thick, black!55,->] (2.5, 1)--(2.5, 0.5);
		
		\draw[thick, black!55, rounded corners=1mm, fill=orange!15, fill opacity=0.5] (4.25, -1.25) -- (4.25, 1.25) -- (4.75, 1.25) -- (4.75, -1.25) -- cycle;
		\node (y1) at(4.5, 1) {$r_1$};
		\node[anchor=mid] (y2) at(4.5, 0) {$\vdots$};
		\node (y3) at(4.5, -1) {$r_L$};
		
		\draw[thick, black!55,->] (3.5, 0)--(4.25, 0.);
		
		\node (y) at(5.75, 0) {FoM ($y$)};
		\draw[thick, black!55,->] (4.75, 0)--(y);	
	\end{tikzpicture}
	\caption{In post-silicon validation, for a modern DUT, there can be up to hundreds tuning conditions ($c_1$ to $c_N$) and tens of tuning knobs ($t_1$ to $t_M$). The resulting high-dimensional data are difficult for human experts to understand and investigate.}
	\label{fig:dut}
\end{figure}
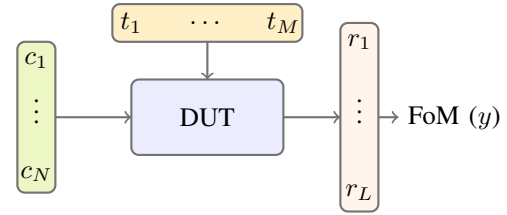

In post-silicon validation, tuning and debugging are typically considered as two major tasks. The former aims to optimally adjust the tuning knobs of manufactured devices in order to meet specifications or maximize the Figure-of-Merit (FoM) defined by manufacturers in order to combat uncertainties (e.g. process variations)~\cite{li2012post,8226046}, while debugging is to identify the root causes for certain defects or flaws in chips. More concretely, given a Device Under Test (DUT) as shown in Fig.~\ref{fig:dut}, we usually record test cases consisting of various tuning conditions $c_1$ to $c_N$ (e.g. process variations or operational conditions) and tuning knobs $t_1$ to $t_M$ with the final FoM $y$ calculated from intermediate test results $r_1$ to $r_L$. Subsequently, domain experts analyze the potential relation between different conditions, tuning knob settings and FoM. This is frequently performed based on the experience of human experts and manual inspection. However, with the development of semiconductor industry, modern devices become increasingly complex and are equipped with tens of tuning knobs. In addition, there can be up to hundreds of diverse tuning conditions. This indicates that a test case of modern chips can be of extremely high dimensionality and thus difficult for human expert to investigate and understand. 

Naturally, with the fast advancements in data-driven approaches and Machine Learning (ML), some previous studies such as~\cite{li2012post,deorio2013machine,7922504,8226046,8429340,7372655,domanski2021self} have developed novel ML-based methods to accelerate post-silicon analysis. Nevertheless, few prior studies have focused on directly reducing data dimensionality in PSV to assist the domain experts in obtaining a deeper and better understanding in test results. Most recently, Liao et. al~\cite{liao2022tuz} has for the first time introduced a variable selection algorithm based on Deep Learning (DL) to PSV data. The basic idea is to use variable selection algorithm to identify the most pivotal candidate variables (e.g. conditions and tuning knobs) that can best predict the FoM. In this way, domain experts need to analyze only the few selected variables instead of high-dimensional raw data. However, these existing approaches were purely data driven, while keeping experts outside the algorithms. This is sometimes risky and less efficient. On one hand, the collected data (e.g. test cases) are typically limited and data-driven approaches without prior knowledge can overfit to the training data, thus losing generalization ability and leading to a threat to reliability. On the other hand, expert knowledge can already provide valuable information (e.g. a few \emph{known} important conditions or tuning knobs) on the obtained data. By incorporating the prior information, users can save time during training. Actually, to the best of knowledge, almost no prior studies towards applying data-driven approaches to test and validation have considered keeping experts in the loop.

Therefore, this paper proposes a novel Conditional Variable Selection (CVS) method based on deep learning, which integrates expert knowledge (i.e. preselected variables by PSV experts) to the training procedure with the intention that our method can reliably identify the most crucial candidate variables to accelerate subsequent post-silicon analysis. Specifically, our method leverages neural networks to encode the preselected variables into latent representations as conditions for training variable selection algorithms. In summary, the main contributions of this work are listed below:
\begin{itemize}
	\item A novel conditional variable selection framework is proposed to take account expert knowledge during training;
	\item Experiments on both synthetic and real-world datasets from the leading manufacturer of automatic test equipment for semiconductors have shown that our method can effectively identify the most critical variables;
	\item Our method has provided a paradigm for other fellow data-driven approaches on integrating expert knowledge into learning algorithms.
\end{itemize}

\section{Related Works}
\label{sec:related_works}
Variable selection has been intensively studied over the last decades in the machine learning community~\cite{guyon2003introduction,li2017feature,dokeroglu2022comprehensive}, which is also known as feature selection or attribute selection. Over the last few years, with the fast development of deep learning, modern variable selection algorithms such as~\cite{gui2019afs,abid2019concrete,9533531} leveraged neural networks to handle large-scale data. Nevertheless, variable selection techniques have not attracted enough attention in the test and semiconductor community. To the best of our knowledge, the most related work is~\cite{liao2022tuz}, in which a DL-based variable selection method was applied to test cases in order to reduce the data dimensionality. However, similar to other existing variable selection approaches, it did not consider prior knowledge as training conditions and therefore kept experts outside the algorithm. 
\section{Methodology}
\label{sec:method}
To enable conditional variable selection with neural networks, we propose to fuse the preselected and candidate variables based on learnable encoding neural layers and feed the fused representations to a neural network targeting a given learning task (e.g. regression or classification). Accordingly, the neural network acts as a guide in a way that the candidate variables, which minimize training losses under the condition of having given the preselected variables, should be assigned with greater importance scores. Thereby, after training, the most critical candidate variables can be easily determined based on the learned scores.

\subsection{Notations}
In this paper, we use the following notations. The input data are denoted as a matrix $X=[\bm{x}_1, \bm{x}_2, \dots, \bm{x}_N]^T\in\mathbb{R}^{N\times D}$, where $N$ is the number of data points (test cases) and $D$ is the number of input variables (tuning conditions and knobs). In the following, we also use $\{v_1, v_2, \dots, v_D\}$ to denote the $D$ different input variables of a given dataset for better readability. According to expert knowledge, $X$ is composed of two parts as $X=[X_p, X_c]$. That is, the \underline{p}reselected $D_p$ variables $X_p = [\bm{x}_{p1}, \bm{x}_{p2}, \dots, \bm{x}_{pN}]^T\in\mathbb{R}^{N\times D_p}$ and the $D_c$ \underline{c}andidate variables $X_c = [\bm{x}_{c1}, \bm{x}_{c2}, \dots, \bm{x}_{cN}]^T\in\mathbb{R}^{N\times D_c}$ with $D = D_c + D_p$. Furthermore, for a supervised conditional variable selection, the labels (FoM or target variables) are denoted as $Y=[\bm{y}_1, \bm{y}_2, \dots, \bm{y}_N]^T\in\mathbb{R}^{N\times D_y}$, where $D_y$ is defined by the given learning task. For example, $D_y=1$ for univariate regression which is one of the most common cases in post-silicon validation.

\subsection{Proposed Framework}
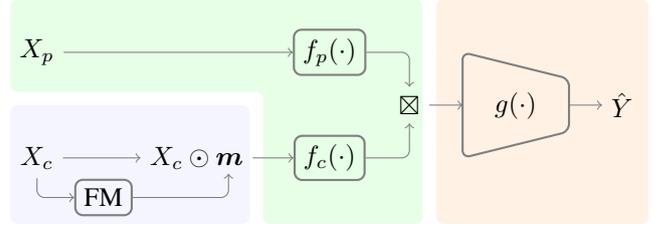
\begin{figure}
	\centering
	\begin{tikzpicture}[scale=0.7]
		\fill[rounded corners=1mm, orange!40, fill opacity=0.25] (7.5, 3) rectangle (11.5, -1.25);
		\fill[rounded corners=1mm, blue!40, fill opacity=0.1] (-0.5, 1) rectangle (4, -1.25);
		\draw[rounded corners=1mm, draw=none, fill=green, fill opacity=0.1] (-0.5, 1.25)--(4.25, 1.25)--(4.25, -1.25)--(7.25,-1.25)--(7.25, 3)--(-0.5, 3)--cycle;

		\node (xc) at(0, 0) {$X_c$};
		\node (xp) at(0, 2) {$X_p$};
		\node (xm) at(3, 0) {$X_c\odot\bm{m}$};
		\node (ocross) at(7, 1) {$\boxtimes$};
		
		\node[thick, rectangle, rounded corners=1mm, draw=gray] (fp) at (5.5, 2) {$f_p(\cdot)$};
		
		\node[thick, rectangle, rounded corners=1mm, draw=gray] (fc) at (5.5, 0) {$f_c(\cdot)$};
		
		\node[thick, rectangle, rounded corners=1mm, draw=gray] (fm) at (1.25, -0.75) {FM};
		
		\draw[thick, rounded corners=1mm, color=gray] (8, 0)--(10, 0.5)--(10, 1.5)--(8, 2.)--cycle;
		\node (g) at(9, 1) {$g(\cdot)$};
		\node (y) at(11, 1) {$\hat{Y}$};
		
		\draw[color=gray, ->] (xc)--(xm);
		\draw[color=gray, ->] (xm)--(fc);
		\draw[color=gray, ->] (xp)--(fp);
		\draw[color=gray, rounded corners=1mm, ->] (xc)--(0, -0.75)--(fm);
		\draw[color=gray, rounded corners=1mm, ->] (fm)--(3.6, -0.75)--(3.6, -0.3);
		
		\draw[color=gray, rounded corners=1mm, ->] (fp)--(7, 2)--(ocross);
		\draw[color=gray, rounded corners=1mm, ->] (fc)--(7, 0)--(ocross);
		\draw[color=gray, ->] (ocross)--(8, 1);
		\draw[color=gray, ->] (10, 1)--(y);
	\end{tikzpicture}
	\caption{The proposed generic framework for conditional variable selection based on neural networks.}
	\label{fig:conditional_fs}
\end{figure}
Based on the notations above, the overall framework is illustrated in Fig.~\ref{fig:conditional_fs}. Broadly speaking, it consists of three major components: \emph{i}) a learnable variable weighting block denoted as FM (\emph{blue}); \emph{ii}) a representation fusion block notated as $\boxtimes$ (\emph{green}); and \emph{iii}) a task-specific neural network $g(\cdot)$ (\emph{orange}).

\subsubsection{Learnable Variable Weighting Block}
This block is designed to learn a vector $\bm{m}=[m_1, m_2, \dots, m_{D_c}]^T\in\mathbb{R}^{D_c}$ with $m_i\in(0, 1)$ which denotes the importance of the corresponding $i$-th candidate variable. The basic idea is inherited from the Feature Mask (FM) method~\cite{9533531} which has been successfully applied to PSV~\cite{liao2022tuz}. The FM-module shown in Fig.~\ref{fig:conditional_fs} can be understood as the following mapping:
\begin{equation}
	\label{eq:fm}
	\bm{m} = \text{FM}(X_c) = \mathit{softmax}\Big(\frac{1}{B}\sum_{i=1}^{B}W_2(W_1\cdot\bm{x}_c + \bm{b}_1) + \bm{b}_2\Big),
\end{equation}
where $W_1\in\mathbb{R}^{L\times D_c}$, $\bm{b}_1\in\mathbb{R}^{L}$, $W_2\in\mathbb{R}^{D_c\times L}$, $\bm{b}_2\in\mathbb{R}^{D_c}$ are trainable parameters of the FM-module and $B$ is the minibatch size during training. The symbol $\odot$ denotes the element-wise multiplication; i.e. $X_c\odot\bm{m} = [\bm{x}_{c1}\odot\bm{m}, \bm{x}_{c2}\odot\bm{m}, \dots, \bm{x}_{cB}\odot\bm{m}]^T\in\mathbb{R}^{B\times D_c}$ during training, corresponding to weighted candidate variables. After training, we obtain the final unique $\bm{m}$ by feeding all training data into the trained FM-module. Then we can easily identify the $k$ most important candidate variables according to the top-$k$ largest entries in $\bm{m}$.

\subsubsection{Representation Fusion Block}
The most conspicuous design in our framework is that we use two learnable non-linear functions $f_p(\cdot)$ and $f_c(\cdot)$ to respectively encode the preselected and candidate variables. Both non-linear functions are implemented by fully connected layers in this work. More precisely, $f_p(\cdot)$ is defined as
\begin{equation}
	\label{eq:encoding}
	f_p(\bm{x}_p) = \sigma(W_p\cdot \bm{x}_p + \bm{b}_p),
\end{equation}
and $f_c(\cdot)$ is defined as
\begin{equation}
	\label{eq:encoding}
	f_c(\bm{x}_c\odot \bm{m}) = \sigma(W_c\cdot (\bm{x}_c\odot \bm{m}) + \bm{b}_c),
\end{equation}
where $W_p\in\mathbb{R}^{L_p\times D_p}$, $\bm{b}_p\in\mathbb{R}^{L_p}$, $W_c\in\mathbb{R}^{L_c\times D_c}$, $\bm{b}_c\in\mathbb{R}^{L_c}$ are trainable network parameters (i.e. weights and biases) and $L_p$ and $L_c$ are user-specified dimensions of the encoded representations. Moreover, $\sigma(\cdot)$ is an activation function to provide non-linearity and $\boxtimes$ denotes the concatenation between the two encoded representations. That is to say, the concatenated representation $\bm{z} = f_p(\bm{x}_p)\boxtimes f_c(\bm{x}_c\odot \bm{m})$ with $\bm{z}\in\mathbb{R}^{L_p+L_c}$.

\subsubsection{Task-Specific Neural Network}
Similar to other DL-based variable selection approaches such as~\cite{gui2019afs,abid2019concrete,9533531}, there are no special requirements on the structure of the task-specific neural network $g(\cdot)$. In this work, $g(\cdot)$ is implemented by a few fully connected layers parameterized by $\bm{\theta}_g$ for simplicity.

\subsection{Learning Objective}
In PSV, we aim to identify which candidate variables can best predict the FoM value which is typically a continuous numerical value. Naturally, this corresponds to a regression task for the entire proposed framework. Therefore, the canonical loss function can be the Mean Squared Error (MSE) loss as
\begin{equation}
	\label{eq:mse}
	\mathcal{L}_\text{MSE} = \frac{1}{N}\sum_{i=1}^{N}||\bm{y}_i-\hat{\bm{y}}_i||_2^2,
\end{equation}
where $||\cdot||_2^2$ denotes the squared 2-norm. Accordingly, the overall training objective is to minimize the MSE loss as
\begin{eqnarray}
	\label{eq:learning_objective}
	\argmin_{\bm\Theta}\frac{1}{N}\sum_{i=1}^{N}||\bm{y}_i - g(f_p(\bm{x}_p)\boxtimes f_c(\bm{x}_c\odot\bm{m}))||_2^2,
\end{eqnarray}
where $\bm{\Theta}$ denotes all trainable parameters of the entire framework (FM, $f_p$, $f_c$ and $g$). As a result, the corresponding training procedure of our framework is summarized in Algorithm~\ref{algo:cvs}.

\begin{algorithm}
	\caption{Conditional Variable Selection}
	\label{algo:cvs}
	\begin{algorithmic}[1]
		\REQUIRE Training dataset pair $(\bm{x}_p, \bm{x}_c, \bm{y})$ in minibatch, learning rate $\alpha$, minibatch size $B$
		\STATE {Randomly initialize the entire framework with the initial neural network parameters $\bm{\Theta}$}
		\FOR {$e=1 $ to maximal training epochs}
		\FOR {$b=1$ to the number of minibatches}
		\STATE{Encode preselected variables as $f_p(\bm{x}_p)$}
		\STATE{Calculate the feature mask on the current minibatch as $\bm{m} = \text{FM}(X_c)$}
		\STATE{Encode the candidate variables as $f_c(\bm{x}_c\odot\bm{m})$}
		\STATE{Concatenate encoded preselected and candidate variables as $\bm{z} = f_p(\bm{x}_p)\boxtimes f_c(\bm{x}_c\odot\bm{m})$}
		\STATE{Calculate current prediction $\hat{\bm{y}} = g(\bm{z})$}
		\STATE{Calculate the loss on the current minibatch as $\mathcal{L}_b = \frac{1}{B}\sum_{i=1}^{B}||\bm{y}_i - \hat{\bm{y}}_i||^2$}
		\STATE{Update network parameters by a gradient decent algorithm as $\bm{\Theta}\leftarrow\bm{\Theta} - \alpha\cdot\nabla_{\bm{\Theta}}\mathcal{L}_b$}
		\ENDFOR
		\ENDFOR
	\end{algorithmic}
\end{algorithm}

\subsection{Implementation}
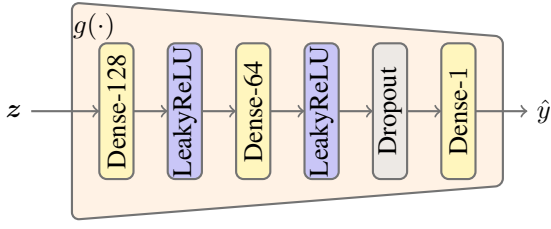
\begin{figure}
	\centering
	\begin{tikzpicture}[scale=0.45]
		\node (x) at(-1.5, 0) {$\bm{z}$};
		
		\draw[thick, draw=black!55, rounded corners=1mm,fill=orange!40,fill opacity=0.25] (0.2, -3.2)--(12.8, -2.2)--(12.8, 2.2)--(0.2, 3.2)--cycle;
		\node (g) at(0.85, 2.5) {$g(\cdot)$};
		
		\draw[thick, draw=black!55, rounded corners=1mm,fill=yellow!40,fill opacity=0.7] (1, -2)--(2, -2)--(2, 2)--(1, 2)--cycle;
		\node [rotate=90] (d1) at(1.5, 0) {Dense-128};
		
		\draw[thick, draw=black!55, rounded corners=1mm,fill=blue!30,fill opacity=0.7] (3, -2)--(4, -2)--(4, 2)--(3, 2)--cycle;
		\node [rotate=90] (d1) at(3.5, 0) {LeakyReLU};
		
		\draw[thick, draw=black!55, rounded corners=1mm,fill=yellow!40,fill opacity=0.7] (5, -2)--(6, -2)--(6, 2)--(5, 2)--cycle;
		\node [rotate=90] (d1) at(5.5, 0) {Dense-64};
		
		\draw[thick, draw=black!55, rounded corners=1mm,fill=blue!30,fill opacity=0.7] (7, -2)--(8, -2)--(8, 2)--(7, 2)--cycle;
		\node [rotate=90] (d1) at(7.5, 0) {LeakyReLU};
		
		\draw[thick, draw=black!55, rounded corners=1mm,fill=lightgray!40,fill opacity=0.7] (9, -2)--(10, -2)--(10, 2)--(9, 2)--cycle;
		\node [rotate=90] (d1) at(9.5, 0) {Dropout};
		
		\draw[thick, draw=black!55, rounded corners=1mm,fill=yellow!40,fill opacity=0.7] (11, -2)--(12, -2)--(12, 2)--(11, 2)--cycle;
		\node [rotate=90] (d1) at(11.5, 0) {Dense-$1$};
		
		\node (y) at(14, 0) {$\hat{y}$};
		
		\draw[thick, black!55, ->] (x) -- (1, 0);
		\draw[thick, black!55, ->] (2, 0) -- (3, 0);
		\draw[thick, black!55, ->] (4, 0) -- (5, 0);
		\draw[thick, black!55, ->] (6, 0) -- (7, 0);
		\draw[thick, black!55, ->] (8, 0) -- (9, 0);
		\draw[thick, black!55, ->] (10, 0) -- (11, 0);
		\draw[thick, black!55, ->] (12, 0) -- (y);
	\end{tikzpicture}
	\caption{The architecture of the task-specific neural network $g(\cdot)$.}
	\label{fig:g}
\end{figure}
In this work, the task-specific network $g(\cdot)$ was implemented with a two-hidden-layer neural network as shown in Fig.~\ref{fig:g}. Both layers were fully connected layers with 128, 64 neurons respectively. LeakyReLU~\cite{maas2013rectifier} with the ratio of 0.02 as
\begin{equation}
\label{eq:leakyrelu}
\text{LeakyReLU}(x)=\left\{
\begin{array}{ll}
		x & \text{ if } x>0\\
		0.02x & \text{ else}
	\end{array} 
\right.
\end{equation}
was used as the activation function for each hidden layer. Before the output layer, we used Dropout~\cite{srivastava2014dropout} with a ratio of 0.3 to avoid overfitting during training. We used Adam optimizer~\cite{kingma2014adam}, which is a modern variant of gradient-descent algorithms, to train our neural networks. All implementations were based on Python and mainly with the TensorFlow framework~\cite{abadi2016tensorflow}.

\section{Motivational Experiments}
\label{sec:exp}
Before diving into real-world datasets, we firstly conducted experiments on synthetic data to justify the effectiveness. In particular, we constructed a synthetic dataset of 2000 training samples with 15 input variables $v_1$ to $v_{15}$. Moreover, we assumed that the regression target $y$ was related to 6 out of 15 input variables based on the formula
\begin{equation}
	\label{eq:synthetic}
	y = v_1^2 + 10v_2 v_3 v_4 + 5v_5 v_6 + \epsilon,
\end{equation}
where $\epsilon\sim N(0, 1)$ and $v_i\sim U(0, 1)$. Obviously, in this synthetic dataset, only the first 6 variables are critical for predicting $y$, while the remaining 9 input variables from $v_7$ to $v_{15}$ are irrelevant to $y$.

In the following subsections, we individually divided the input variables into preselected and candidate variables to investigate whether our approach can learn correct importance scores for the candidate variables given certain preselected variables as training condition.
\begin{table*}[!h]
	\renewcommand{\arraystretch}{1.2}
	\centering
	\caption{Exemplary (anonymous) test cases of the real-world dataset.}
	\label{tab:datasets}
		\begin{tabular}{c|c|c|c|c|c|c|c|c|c|c|c|c|c|c}
			\hline
			& $v_1$ & $v_2$ & $v_3$ & $v_4$ & $v_5$ & $v_6$ & $v_7$ & $v_8$ & $v_9$ & $v_{10}$ & $v_{11}$ & $v_{12}$ & $v_{13}$ & $y$ \\
			\hline
			$\bm{x}_1$ &3&1&0.22&1.34&0.54&2.44&4&12&0&0.98&2&6&1&3.12\\
			\hline
			$\vdots$ & \multicolumn{12}{c}{$\cdots$} && $\vdots$\\
			\hline
			$\bm{x}_{900k}$ &8&3&1.71&0.99&1.10&2.67&7&2&4&1.13&3&2&2&-1.8\\
			\hline
		\end{tabular}
\end{table*}
\subsection{Experiment I: Single Preselected Variable as Condition}
\begin{figure}
	\centering
	\includegraphics[width=.9\linewidth]{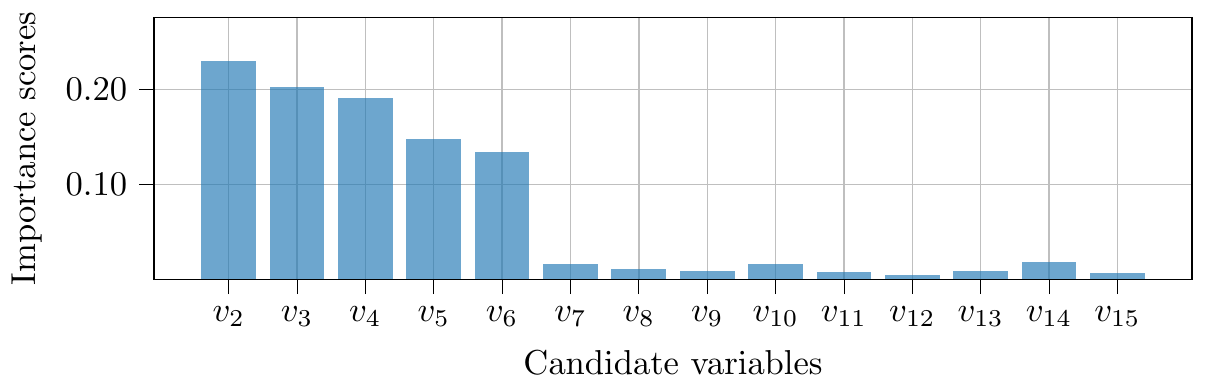}
	\caption{$v_1$ was preselected as training condition.}
	\label{fig:FS_Synthetic_x0}
\end{figure}

\begin{figure}
	\centering
	\includegraphics[width=.9\linewidth]{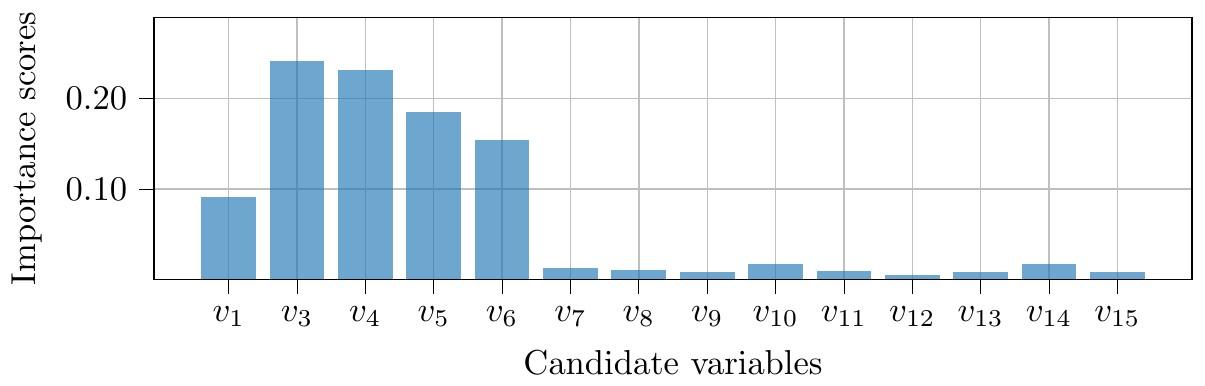}
	\caption{$v_3$ was preselected as training condition.}
	\label{fig:FS_Synthetic_x2}
\end{figure}

Fig.~\ref{fig:FS_Synthetic_x0} shows the learned variable importance given the preselected variable $v_1$ as the training condition. It can be easily seen that $v_2$ to $v_6$ were successfully assigned with large importance scores after training, while the remaining nine variables were with extremely small scores. This matches the relation defined in Eq.~\ref{eq:synthetic}. In a separate experiment, we considered $v_2$ as the preselected variable as training condition. The resulting learned scores are presented in Fi.~\ref{fig:FS_Synthetic_x2}. As expected, the irrelevant variables from $v_7$ to $v_{15}$ were still assigned with small values in comparison to the relevant variables ($v_1$, $v_3$ to $v_6$). Another inspiring observation is that the learned importance scores really reflects the relative contribution of each individual candidate variables, meaning that $v_3$ and $v_4$ were more important under the condition of $v_2$, while $v_1$ had the smallest importance among all relevant candidate variables. 

\subsection{Experiment II: Multiple Preselected Variables as Condition}
\begin{figure}
	\centering
	\includegraphics[width=.9\linewidth]{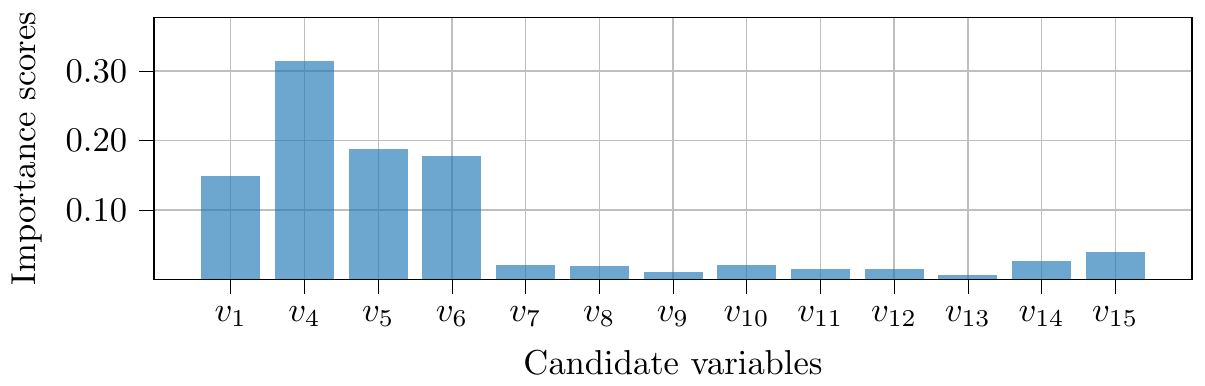}
	\caption{$v_2$ and $v_3$ were preselected as training condition.}
	\label{fig:FS_Synthetic_x1_x2}
\end{figure}

An important advantage of our approach is that training conditions can be multiple preselected variables. To justify this property, we conducted an experiment by simultaneously preselecting both $v_2$ and $v_3$ as training conditions and the learned importance scores are shown in Fig.~\ref{fig:FS_Synthetic_x1_x2}. Obviously, in this case, $v_4$ was assigned with significantly larger importance scores than other relevant candidate variables because $v_4$ is more important, having given $v_2$ and $v_3$ as conditions.

\subsection{Experiment III: Conditional Variable Selection with Redundant Variables}
\begin{figure}
	\centering
	\includegraphics[width=.9\linewidth]{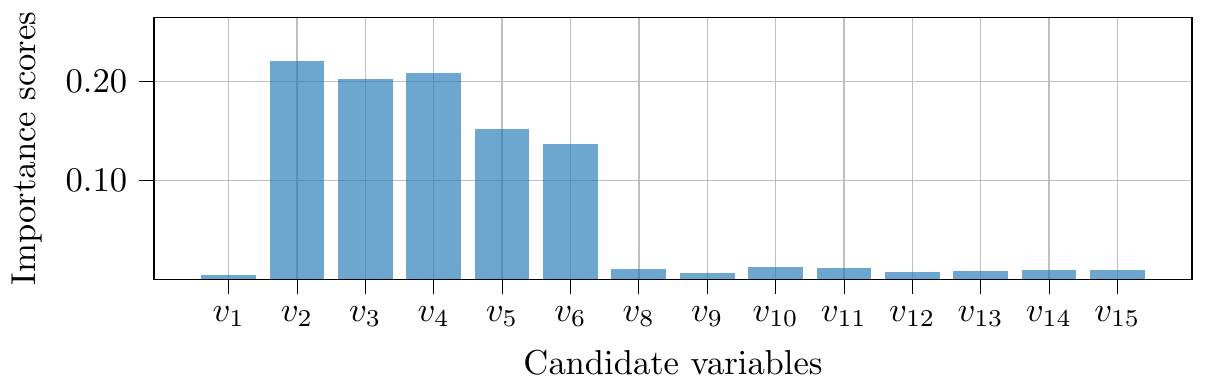}
	\caption{$v_7$ was redundant to $v_1$ and preselected.}
	\label{fig:FS_Synthetic_x6_r}
\end{figure}

Avoiding selecting redundant variables is a challenging task in variable selection~\cite{guyon2003introduction}. In order to study how our algorithm behaves in the presence of redundant variables, we additionally let the input variable $v_7 = v_1^2$ be a redundant variable towards $v_1$, and $v_7$ was the preselected variable as training condition. From Fig.~\ref{fig:FS_Synthetic_x6_r}, we can easily see that $v_2$ to $v_6$ were successfully assigned with large importance and thus selected, while the redundant variable $v_1$ was with an extremely small score. This means that our method can automatically avoid selecting the candidate variables which are redundant to the conditions (preselected variables). This property also suggests that the condition has been successfully involved into the training procedure.


\section{Case Study on Real-World Data}
\label{sec:case}
This section studies the conditional variable selection performance of the proposed framework on a real-world large-scale PSV dataset.

\subsection{Datasets}
The real-world PSV dataset is provided by a leading manufacturer of automatic test equipment for semiconductor industry. Specifically, the dataset contains test cases from 9 different DUTs, in which each test case is composed of 13 input variables and one calculated FoM as target variable. It should be emphasized that the input variables of this real-world dataset are of mixed data types, meaning that there are both numerical (continuous and discrete) and categorical variables. Although mixed data types can be a serious challenge to conventional variable selection approaches, our method based on neural networks can well handle mixed data types. For each single DUT, we collected 100,000 test cases and there are in total 900,000 test cases (training samples). Additionally, due to the confidentiality, in this section we omit the concrete physical meaning for individual variables and only use $v_1$ to $v_{13}$ to represent them. To provide a more intuitive impression on this dataset, TABLE~\ref{tab:datasets} shows the exemplary data in an anonymous way that all quantitative values are artificial to maintain the confidentiality of the data for the company.

\subsection{Case Study: CVS on Single DUT}
\begin{figure}
	\centering
	\includegraphics[width=.89\linewidth]{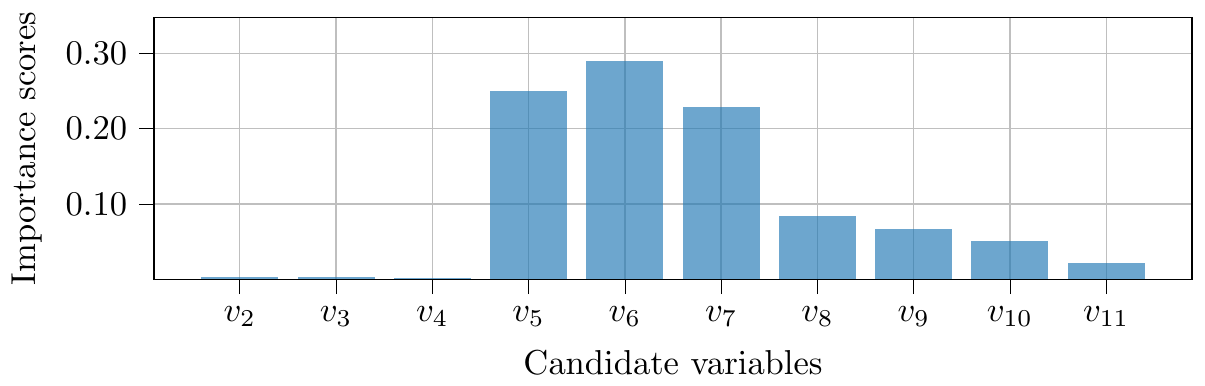}
	\caption{$v_1$ as condition on a single DUT.}
	\label{fig:FS_dev0_c0}
\end{figure}

\begin{figure}
	\centering
	\includegraphics[width=.89\linewidth]{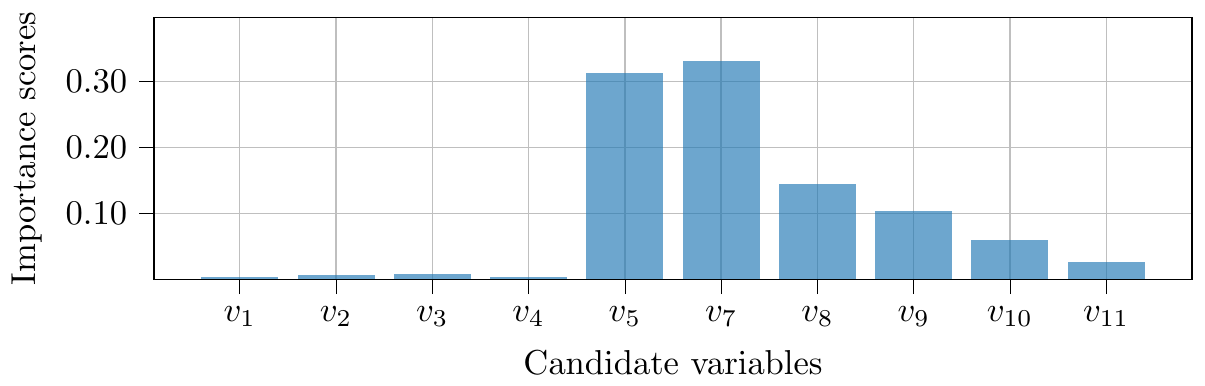}
	\caption{$v_6$ as condition on a single DUT.}
	\label{fig:FS_dev0_t2}
\end{figure}

\begin{figure}
	\centering
	\includegraphics[width=.89\linewidth]{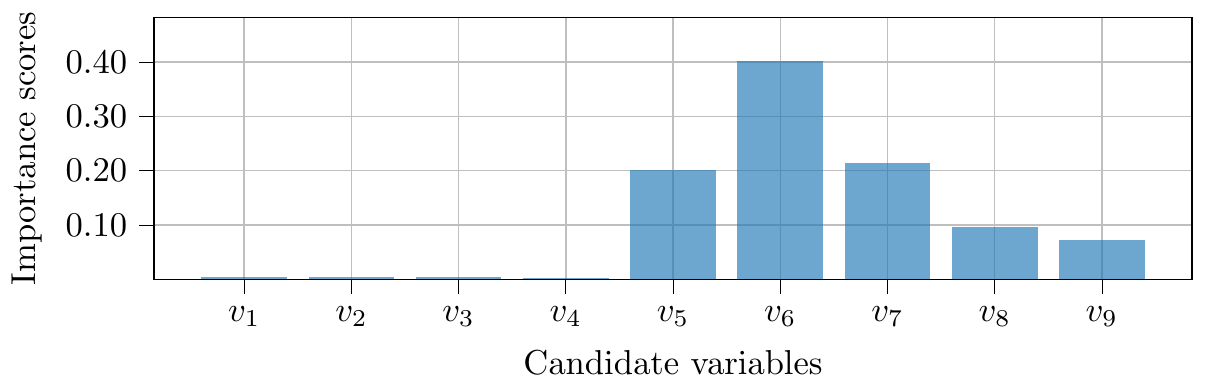}
	\caption{$v_{10}$ and $v_{11}$ as condition on a single DUT.}
	\label{fig:FS_dev0_t6_t7}
\end{figure}

We first considered one single DUT only. It should be noted that two input variables were removed before training on a single DUT in this subsection, because they were identical within a given DUT, meaning that we considered 11 variables in total. Fig.~\ref{fig:FS_dev0_c0} and Fig.~\ref{fig:FS_dev0_t2} show the learned importance scores by preselecting $v_1$ and $v_6$ as conditions, respectively. Apparently, given $v_1$ as condition, $v_5$ to $v_{11}$ were all important, while $v_5$ to $v_7$ were significantly more critical than other variables. Given $v_6$ as conditions, $v_5$ and $v_7$ were the most important candidate variables, while $v_1$ seemed to be irrelevant in this case, which matched the exhaustive search. Furthermore, we simultaneously let $v_{10}$ and $v_{11}$ be the preselected variables as the training condition. Fig.~\ref{fig:FS_dev0_t6_t7} presents the importance scores. Interestingly, we can obviously observe that $v_6$ is notably more important than other candidate variables, taking $v_{10}$ and $v_{11}$ into consideration. 

It should be emphasized that an exhaustive search can take significantly longer time than our algorithm. For example, considering a case where we want to identify the most critical five variables given one preselected variable as condition, an exhaustive search requires to explore $\binom{10}{5}=252$ candidate variable combinations for a single DUT. This implies that we need to train a model for 252 times to obtain the final selection results, while our method can provide the importance scores for training only once.


\subsection{Case Study: CVS on Multiple DUT}

\begin{figure}
	\centering
	\includegraphics[width=.89\linewidth]{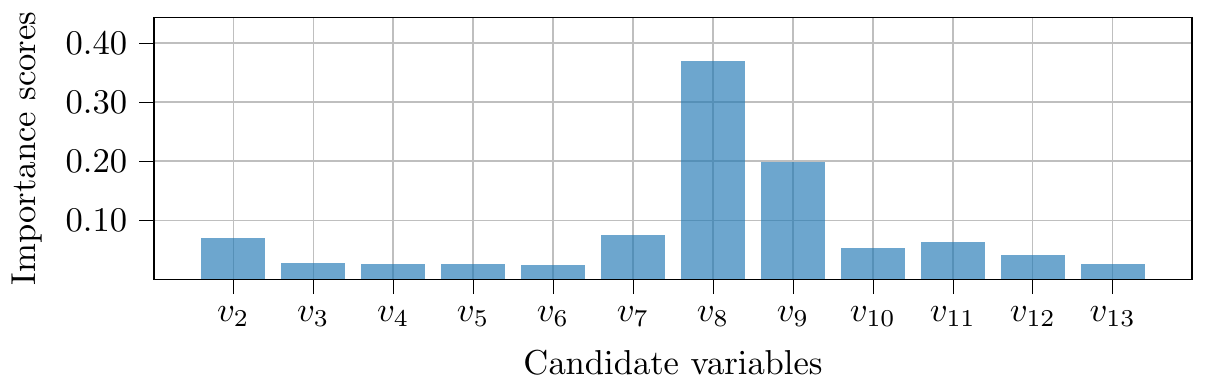}
	\caption{$v_1$ as condition on all nine DUTs.}
	\label{fig:FS_all_dev}
\end{figure}

In this experiment, we considered all 900k test cases. As an example, $v_1$ was preselected as training condition and Fig.~\ref{fig:FS_all_dev} shows the learned importance scores. We can observe that $v_2$ and $v_7$ to $v_{11}$ were notably more critical than the other candidate variables under the condition of preselecting $v_1$, which matched our exhaustive search results.

\begin{figure}
	\centering
	\includegraphics[width=.89\linewidth]{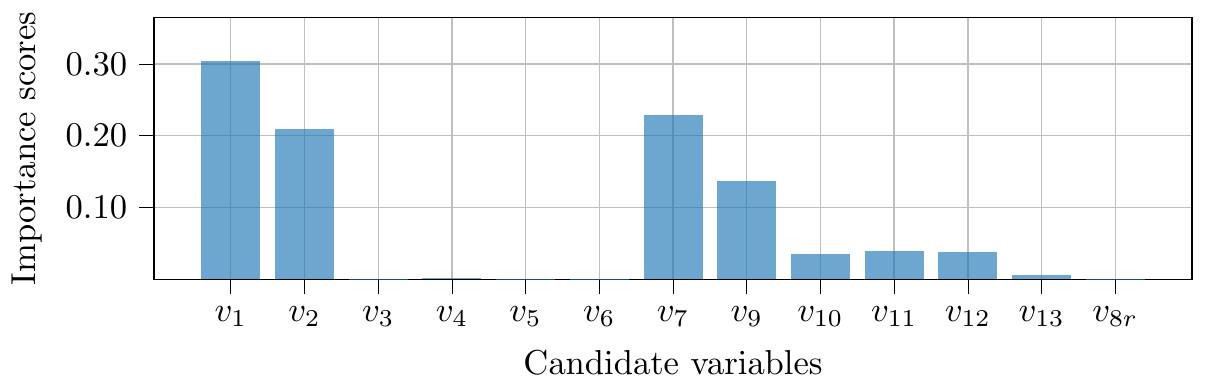}
	\caption{$v_8$ as condition on all nine DUTs and a redundant variable $v_{8r}$ was added to the candidate variable list.}
	\label{fig:FS_all_t2_r}
\end{figure}

Next, we justified the effectiveness of our framework under redundant candidate variables. In particular, we constructed a redundant candidate variable $v_{8r}$ towards $v_8$ by a simple duplication as $v_{8r}=v_8$. The resulting importance scores are shown in Fig.~\ref{fig:FS_all_t2_r}. As expected, $v_{8r}$ was assigned with extremely small score and not selected by our algorithm. 

\begin{figure}
	\centering
	\includegraphics[width=.82\linewidth]{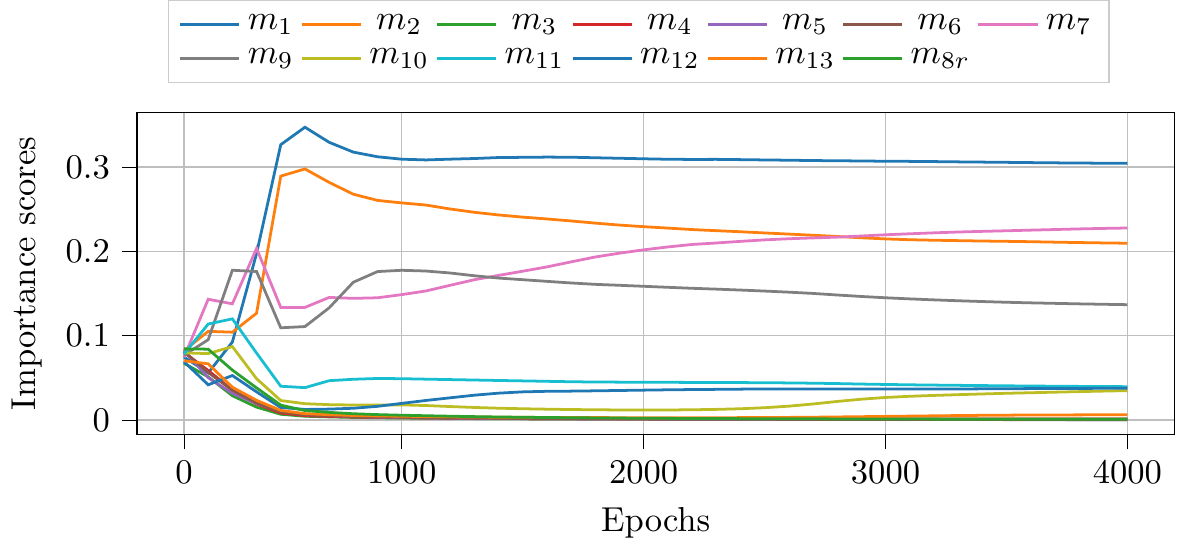}
	\caption{The learned feature mask $\bm{m}$ during training.}
	\label{fig:mask_learning}
\end{figure}

Furthermore, Fig.~\ref{fig:mask_learning} shows the values of each entry in the variable importance vector $\bm{m}$ during training. We can see that all values were similar at the beginning, because we intentionally initialized the network in such a way that all candidate variables were equally considered to avoid bias. Obviously, the values of all entries in $\bm{m}$ converged in around 4k epochs, corresponding to about 5 minutes on a consumer graphics processing unit on desktop computer.

\subsection{Unexpected Discovery}
Fig.~\ref{fig:FS_all_t2_r} shows that our algorithm assigned similar scores to both $v_1$ and $v_2$, which surprised the domain experts of the anonymous company. According to the experts' experience, $v_1$ and $v_2$ should be redundant to each other. However, our algorithm assigned apparently large importance scores to both variables. In addition, in Fig.~\ref{fig:FS_all_dev}, $v_2$ was also important under the condition of preselecting $v_1$, seemingly contradicting to the previous experiments for redundant variables. 

This observation suggests the conflicts between data-driven approaches and expert knowledge, motivating the domain experts to rethink the relation between $v_1$ and $v_2$. Coincidentally, in this dataset, there exists a relation between both variables as
\begin{equation}
	\label{eq:dev_cor}
	v_2=\left\{
	\begin{array}{ll}
		1, & \text{if } v_1\in\{1, 2, 3\}\\
		2, & \text{if } v_1\in\{4, 5, 6\}\\
		3, & \text{if } v_1\in\{7, 8, 9\}
	\end{array} 
	\right.
\end{equation}
This suggests that $v_1$ and $v_2$ might have provided information of different levels during training. In other words, $v_1$ provided information for considering test cases into 9 groups, while $v_2$ provided information for categorizing test cases into 3 groups.
\section{Discussion}
\label{sec:discussion}
The conducted experiments presented in the previous section have shown how our conditional variable selection algorithm assists expert to efficiently identify the most critical candidate variables under the condition of preselecting a few variables. Actually, the potential of our proposed idea is much more than this. Firstly, as shown above, the preselected variables were considered as expert knowledge to guide the training procedure. However, we suggest that almost other meta information in addition to expert knowledge can be used in our framework by considering them as $X_p$. For example, manufacturing dates and locations, process parameters and chip specifications can be fed to our framework as conditions for training. Moreover, the idea of keeping expert in the loop can be used to other related fields such as wafer map defect pattern classification, where many recent studies~\cite{liao2022wafer} have focused on using deep neural networks to perform classification. In this case, some expert knowledge such as wafer test lots or historical test information can be fused and fed to the original neural network to further enhance the efficiency and effectiveness.
\section{Conclusion}
\label{sec:conclusion}
This paper proposed a novel framework of conditional variable selection for efficient post-silicon analysis by taking expert knowledge into account during training. Our approach was based on artificial neural networks and can thus easily handle large-scale data of high-dimensionality. Experiments on synthetic datasets clearly presented the effectiveness of our framework under different training conditions. Moreover, the experiments on real-world datasets confirmed its superiority in practice. Last but not least, this work is expected to inspire other fellow researchers to consider expert knowledge into data-driven approaches to enhance the overall efficiency, reliability and performance.


\bibliographystyle{IEEEtran}
\bibliography{Refs}

\end{document}